\documentclass[letterpaper, 10 pt, conference]{ieeeconf}
\IEEEoverridecommandlockouts

\usepackage{graphicx}
\usepackage{amsmath,amssymb}
\usepackage{booktabs}
\usepackage{xcolor}
\usepackage{array}
\usepackage{multirow}
\usepackage{multicol}
\usepackage{tabularx}
\usepackage{adjustbox}
\usepackage{float}
\usepackage{caption}
\usepackage{subcaption}
\usepackage{balance}
\usepackage{stfloats} % for figure* placement control (strip, etc.)
\usepackage[hidelinks]{hyperref} % MUST be loaded for \href and \autoref
\usepackage{makecell}

\IEEEaftertitletext{%
  \vspace*{-0.8\baselineskip} % tighten gap under the title
  \begin{center}
    \includegraphics[width=\textwidth]{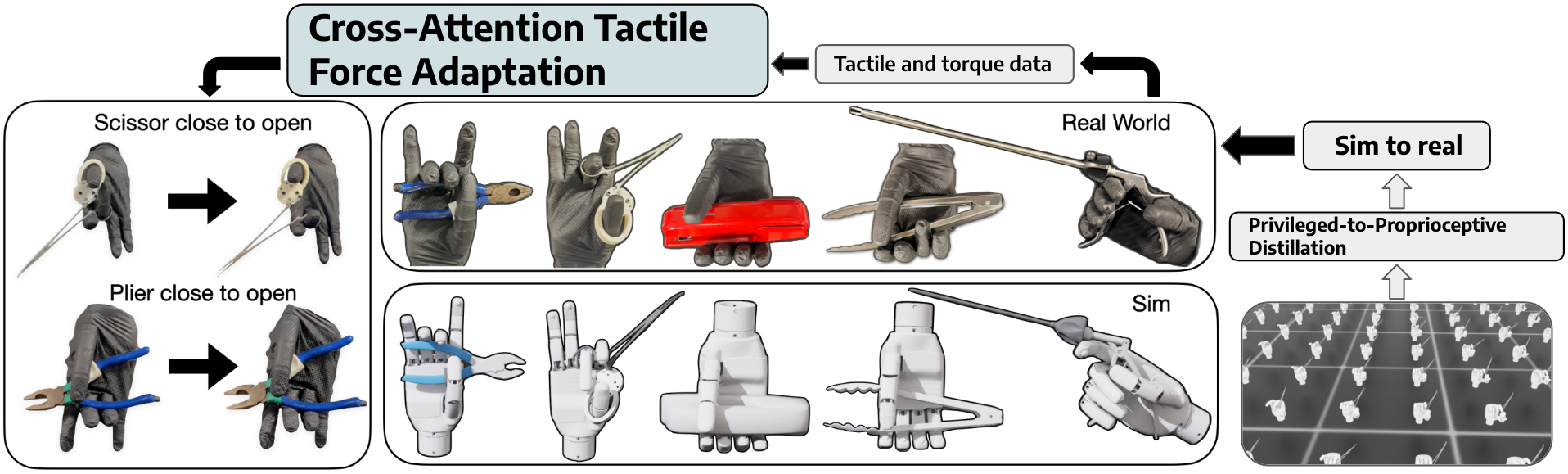}
    \captionof{figure}{An illustration is shown where a privileged oracle is trained in simulation and distilled into a proprioceptive student policy for hardware deployment. Real-world tactile and motor torque signals are then used to train the CATFA module, which refines the transferred policy through cross-attention for contact-aware adaptation.}
    \label{fig:cover_photo}
  \end{center}
}

\begin{document}
\title{\LARGE \bf In-Hand Manipulation of Articulated Tools with Dexterous Robot Hands with Sim-to-Real Transfer}

% Authors Redacted
% \author{Anonymous Authors}
\author{Soofiyan Atar$^{1}$, Daniel Huang$^{1}$, Florian Richter$^{1}$,  Michael Yip$^{1}$, \IEEEmembership{Senior Member, IEEE}
% %\thanks{*This work was not supported by any organization}% <-this % stops a space
\thanks{$^1$Electrical and Computer Engineering Department, University of California, San Diego, La Jolla, CA 92093 USA. {\tt\footnotesize\{satar,  yah032, frichter, yip\}@ucsd.edu}}
}

\maketitle
 
% \begin{figure*}[!t] 
%     \centering 
%     \includegraphics[height=0.6\columnwidth]{images/cover_photo_paper.png} 
%     \caption{Cover photo} 
%     \label{fig:cover_photo} 
% \end{figure*}

\begin{abstract}

Reinforcement learning (RL) and sim-to-real transfer have advanced rigid-object manipulation. However, policies remain brittle for \texttt{articulated} mechanisms due to contact-rich dynamics that require both stable grasping and simultaneous free in-hand articulation. Furthermore, articulated objects and robot hands exhibit under-modeled joint phenomena such as friction, stiction, and backlash in real life that can increase the sim-to-real gap, and robot hands still fall short of idealized tactile sensing, both in terms of coverage, sensitivity, and specificity. In this paper, we present an original approach to learning dexterous in-hand manipulation of articulated tools that has reduced articulation and kinematic redundancy relative to the human hand. Our approach augments a simulation-trained base policy with a sensor-driven refinement learned from hardware demonstrations. This refinement conditions on proprioception and target articulation states while fusing whole-hand tactile and force–torque feedback with the policy’s action intent through cross-attention. The resulting controller adapts online to instance-specific articulation properties, stabilizes contact interactions, and regulates internal forces under perturbations. We validate our method across diverse real-world tools, including scissors, pliers, minimally invasive surgical instruments, and staplers, demonstrating robust sim-to-real transfer, improved disturbance resilience, and generalization across structurally related articulated tools without precise physical modeling. \href{https://sites.google.com/view/in-hand-manipulation-of-artic/home}{Website}.
\end{abstract}

\vspace{-2mm}
\section{Introduction}

A central goal of robotics is to enable operation in human-centric environments \cite{atar2025humanoidshospitalstechnicalstudy}, which require interacting with tools designed for human hands. While manipulation of rigid objects has progressed significantly \cite{qi2025simplecomplexskillscase, wang2024lessonslearningspinpens}, \textbf{articulated tools} with internal kinematics, such as scissors or pliers, remain fundamentally challenging. Their joint-dependent dynamics and contact constraints demand precise coordination beyond standard rigid-object control. This challenge is especially critical for humanoid robots \cite{6630792, Liu_2024}, which, with their dexterous multi-fingered hands, are expected to use tools and operate in environments built for people \cite{werby2025articulatedobjectestimationwild, 10204320}. Advancing dexterous in-hand manipulation of articulated tools is therefore a key step toward scalable real-world deployment and practical humanoid capability \cite{ASFOUR200854}.

Beyond sim-to-real transfer, other learning-based strategies have been explored for dexterous manipulation. Imitation learning \cite{lin2024learningvisuotactileskillsmultifingered, li2024okamiteachinghumanoidrobots} and foundation models \cite{zhang2025adaptivearticulatedobjectmanipulation} leverage human demonstrations to guide policy training. Demonstration data are often collected via teleoperation with dexterous hands, sometimes enhanced with haptic feedback \cite{lin2025typetelereleasingdexterityteleoperation}. While effective, these approaches face limited scalability due to the high cost and time required for demonstration collection. Another direction is hierarchical policy learning and task planning \cite{5980391, Barto2003RecentAI}, which decomposes complex skills into structured sub-tasks. However, when low-level skills fail to provide accurate or robust feedback, minor execution errors cascade upward, ultimately degrading the performance of higher-level plans. These limitations become particularly pronounced in articulated, contact-rich settings, where precise force regulation and internal joint coordination are critical.

The prevailing paradigm for training dexterous hands relies on sim-to-real policy transfer \cite{wang2024lessonslearningspinpens, qi2022inhandobjectrotationrapid}. While effective for rigid objects and suction-based grasping \cite{atar2024optigraspoptimizedgrasppose, yang2023dynamograsp}, this paradigm breaks down in a dynamic, contact-rich manipulation—particularly for articulated tools held in high-degree-of-freedom (DOF) robot hands. Internal joint coupling, friction, stiction, and contact constraints are difficult to model, thereby amplifying reality gaps and degrading policy reliability. Articulated tools further introduce joint-dependent dynamics and nonlinear contact interactions, widening this gap. Although tactile and visuo-tactile feedback can partially mitigate state uncertainty \cite{Wang_2022, qi2023generalinhandobjectrotation}, current simulators remain fundamentally limited in accurately modeling multi-body contact under complex dynamical and kinematic constraints, leading to brittle policies and poor transfer \cite{acosta2022validatingroboticssimulatorsrealworld}.
Taken together, these limitations reveal a fundamental gap; current learning-based approaches remain brittle when confronted with the coupled kinematics and contact dynamics of articulated tools in the real world.

In this work, we demonstrate that dexterous robotic hands can reliably manipulate articulated tools under real-world contact dynamics, without relying on perfectly modeled simulation or large-scale teleoperated demonstrations.
To achieve this, we present the following contributions:
\begin{itemize}
    \item A disturbance-driven sim-to-real training pipeline that distills a privileged simulation policy into a student policy trained with structured force–torque random-walk perturbations to improve contact robustness.
    \item Cross-Attention Tactile Force Adaptation (CATFA), an intent-conditioned adaptation module that fuses the frozen base policy embedding with real tactile and force–torque feedback via multi-head cross-attention to produce contact-aware actions.
    \item Comprehensive real-world evaluation across five articulated tools, including perturbation analysis and quantitative robustness benchmarking.
\end{itemize}
We validate our approach both in simulation and on a physical Franka arm with a dexterous hand, showing substantial improvements in grasp stability and disturbance robustness across all five articulated tools.

\section{RELATED WORKS}

In-hand manipulation \cite{probabilisticroadmaps, trinklerolling, doi:10.1177/0278364919887447} has seen significant progress through reinforcement learning (RL) within a sim-to-real framework. Seminal works \cite{openai2019solvingrubikscuberobot} in reorienting a rigid cube and demonstrating dynamic pen spinning have shown that complex skills can be learned entirely in simulation. These methods typically rely on extensive domain randomization to train policies that are robust enough for zero-shot or open-loop transfer to the real world \cite{Bhatt__2021, patidar2023inhandcubereconfigurationsimplified}. However, this approach often proves brittle, as the policies still struggle to overcome the reality gap caused by unmodeled, contact-rich physics.

Recent work shows that input representations such as point clouds can significantly improve generalization to novel objects with multi-fingered hands, enabling sim-to-real transfer with minimal real-world tuning \cite{qin2022dexpointgeneralizablepointcloud}. The benefit of rich tactile feedback \cite{yang2024anyrotategravityinvariantinhandobject} has been demonstrated, with zero-shot transfer of rotational manipulation tasks for unseen objects under varying contact conditions.

Separately, a significant body of research has focused on the broader challenge of articulated-object manipulation, with an emphasis on tasks such as opening doors, drawers, and cabinets. Prominent works \cite{wang2025articubotlearninguniversalarticulated, Xu_2022} have successfully trained single, general-purpose policies that map visual inputs (typically point clouds) to robot arm actions, enabling zero-shot transfer to a variety of unseen objects. These methods have made great strides in solving the perception and motion planning problems associated with interacting with articulated mechanisms in the environment. However, these approaches treat the articulated object as part of the static world and do not address the distinct and more dynamic challenge of in-hand manipulation, where the object is held and controlled entirely within the grasp of a dexterous hand, thereby requiring the palm and fingers to achieve both objectives of stable grasp, but also articulability of the object.

The challenge of in-hand manipulation is amplified for articulated tools, whose internal dynamics are complex to simulate. Simple cases, such as power tools with a single trigger finger, allow grasp and articulation to be decoupled. Still, tools like scissors or laparoscopic instruments involve coupled motions that create complex contact interactions. Recent work on tweezer manipulation~\cite{xu2025hierarchicalreinforcementlearningarticulated} has shown progress, but remains limited to a single tool and relies on privilege-informed controllers, reducing generalizability and bypassing the challenge of online skill discovery. In contrast, our approach introduces a post-transfer adaptation mechanism that learns robust, generalizable policies directly from real-world interaction.

Because of the complexities of in-hand manipulation of articulated tools and the limited observations available from in-hand cameras, the importance of tactile feedback \cite{jawale2024tactile} quickly becomes important. However, far fewer recent works explore non-vision-based tactile skin simulation that realistically captures both normal and shear forces. Some simulate simplified skin deformation \cite{9123915, 9158822} and sensor response handling, cross-talk, and noise, but remain intractable for training reinforcement learning policies. Others use binary or coarse tactile sensors (e.g., force-sensitive resistors), and other simulators do not represent shear dynamics under contact \cite{yuan2024robotsynesthesiainhandmanipulation, yin2023rotatingseeinginhanddexterity}. In contrast, our work employs whole-hand tactile sensing, providing quantized continuous force readings, which are structured into a tactile image and embedded before cross-attention conditioning of the BC policy.

% Hand: $(q_t, \dot{q}_t, u_t)$

% Articulation: $(\theta^{\text{art}}_t, \dot{\theta}^{\text{art}}t, \theta^{\text{target}}{s_t})$

% Object: $(x^{\text{obj}}_t, \dot{x}^{\text{obj}}_t, q^{\text{obj}}_t, h^{\text{obj}}_t)$

% Contacts: $(n_t, n^\star)$

% Forces: $\tau^{\text{raw}}_t$

\section{Methods}

\begin{figure*}[t]
    \centering
    % Left image
    \begin{subfigure}[b]{0.69\textwidth}
        \centering
        \includegraphics[width=\textwidth]{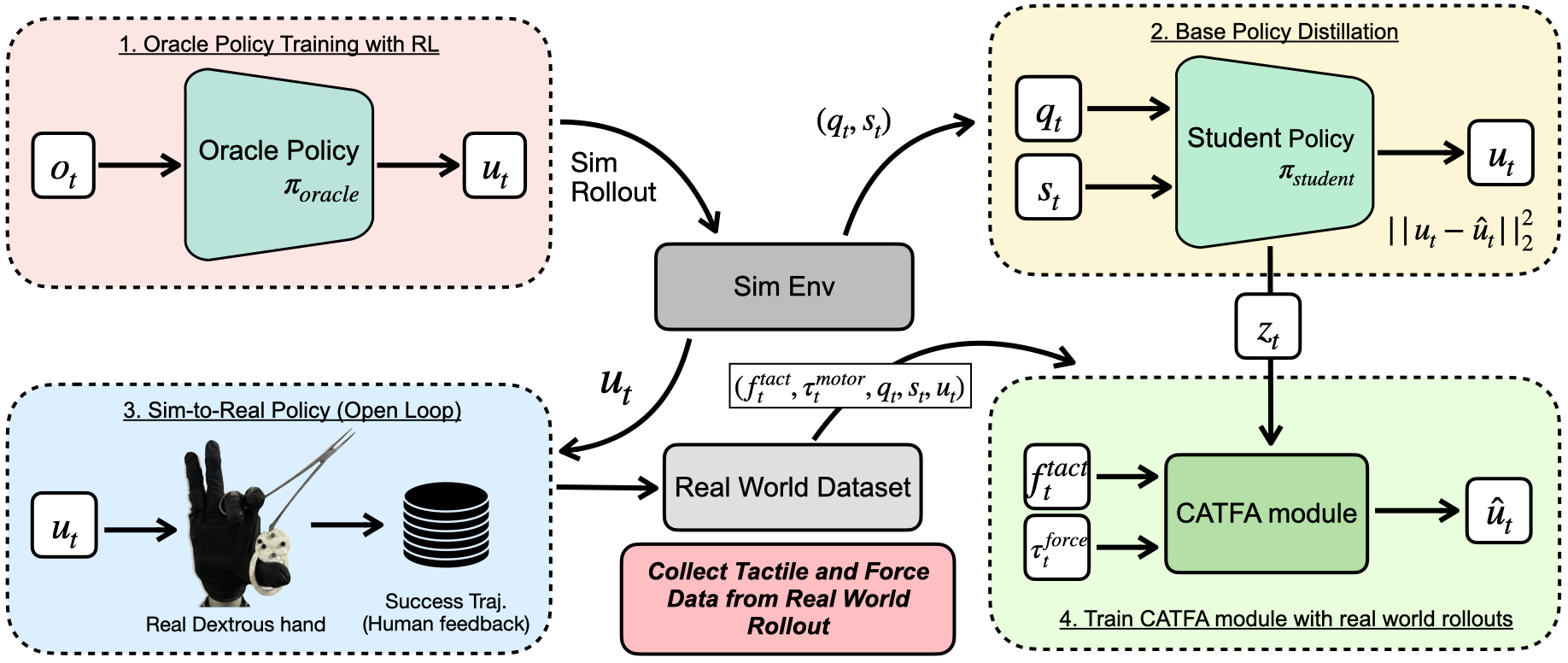}
        \caption{}
        \label{fig:overview-left}
    \end{subfigure}
    \hfill
    % Right image with vertical adjustment
    \begin{subfigure}[b]{0.29\textwidth}
        \centering
        \includegraphics[width=\textwidth]{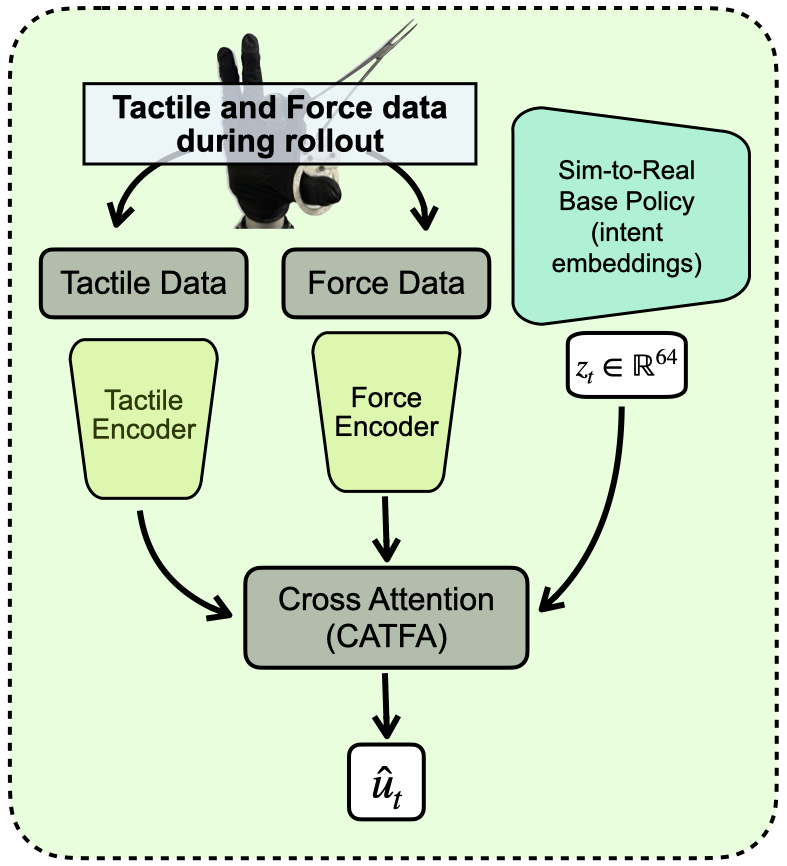}
        \caption{}
        \label{fig:catfa}
    \end{subfigure}
    \caption{
        Technical overview of our pipeline for articulated in-hand manipulation. (a) Oracle training with privileged observations in simulation, followed by distillation into a proprioceptive base policy and real-world data collection with additional tactile $f^{\text{tact}}_t$ and motor torque $\tau^{\text{motor}}_t$ signals. (b) CATFA fuses these sensory modalities with policy intent via cross-attention to enable online adaptation on hardware. This design establishes a disturbance-aware sim-to-real framework that refines contact interactions and improves robustness under articulation-dependent dynamics.}
    \label{fig:overview}
    \vspace{-5mm}
\end{figure*}

% \begin{figure*}
%     \centering
%     \includegraphics[width=0.75\textwidth]{images/pipeline.png}
%     \caption{Sim to real transfer using open loop and then using the CATFA module in the real world for online adaptation}
%     \label{fig:overview}
% \end{figure*}

Our framework, Fig.~\ref{fig:overview}, consists of three stages: (1) training a privileged oracle policy in simulation with disturbance augmentation, (2) distilling the oracle into a deployable proprioceptive base policy for sim-to-real transfer, and (3) enabling online adaptation on hardware via a cross-attention tactile–force refinement module (CATFA). Each stage addresses a specific limitation of articulated in-hand manipulation, progressively bridging the gap between simulation robustness and real-world contact uncertainty.

\subsection{Oracle Policy Definition and Training}
\label{sec:oracle_policy}
% In our setting, an oracle policy $\pi_{oracle}$ is required because the dynamics of articulated tools are difficult to explore reliably through random initialization or teleoperation.
We first describe the simulation stage in which a privileged oracle policy is trained to achieve stable articulation behavior under structured perturbations. We define an oracle policy, $\pi_{\text{oracle}}$, with privileged observations to learn a base controller for in-hand articulated manipulation. The oracle is trained in simulation using PPO~\cite{schulman2017proximalpolicyoptimizationalgorithms} with a curriculum of force–torque perturbations applied to the articulated tool. The disturbance magnitude is gradually increased to emulate arbitrary gravity vectors and external contact dynamics, stabilize articulation under simulated disturbances (see Sec.~\ref{sec:perturbations}). At each timestep $t$, the oracle outputs absolute joint targets $u_t \leftarrow \pi_{\text{oracle}}(o_t)$ from privileged observations $o_t$. The privileged signals are later removed via distillation for sim-to-real transfer, which is more efficient than directly training with non-privileged observations.

\noindent\textbf{Initial State.} The articulated tool is initialized using teleoperated demonstrations with a hand tracker (Manus gloves~\cite{manus2025}) when a stable grasp cannot be achieved autonomously. This ensures rollouts begin from feasible hand–tool configurations despite articulation variability.

\noindent\textbf{Observations.} The privileged observation at time $t$ is
\[
o_t = \big[\, q_t, \dot{q}_t, u_{t-1}, \theta^{\text{art}}_t, \dot{\theta}^{\text{art}}_t, x^{\text{obj}}_t, \dot{x}^{\text{obj}}_t, s_t, \tau^{\text{raw}}_t \,\big],
\]
where $q_t \in \mathbb{R}^{6}$ and $\dot{q}_t \in \mathbb{R}^{6}$ are joint positions and velocities of the active DoFs (four finger joints and two thumb joints of the Inspire hand), and $u_{t-1} \in \mathbb{R}^{6}$ is the previous joint target. The articulation state is given by $\theta^{\text{art}}_t \in \mathbb{R}$ and $\dot{\theta}^{\text{art}}_t \in \mathbb{R}$. The tool pose is represented by $x^{\text{obj}}_t \in \mathbb{R}^3$ and $\dot{x}^{\text{obj}}_t \in \mathbb{R}^3$, corresponding to the position and linear velocity of a reference frame attached to the articulation joint axis (Fig.~\ref{fig:object_set}), expressed in the robot base frame. This frame is defined in the URDF and does not correspond to the center of mass or tool tip. Raw simulated joint force signals are $\tau_t^{\text{raw}} \in \mathbb{R}^6$. The binary one-hot articulation command $s_t \in \{0,1\}^2$ specifies the desired state (e.g., open or closed). The previous action $u_{t-1}$ is included only in the oracle’s privileged observation to compensate for action smoothing and accumulated perturbations, and is removed during distillation.

\noindent\textbf{Actions.} At each timestep, the policy outputs absolute joint targets $u_t \in \mathbb{R}^{6}$ for the hand. To improve stability and reduce vibration, the executed command is smoothed using an exponential moving average:
\begin{equation}
\tilde{u}_t \leftarrow \alpha u_t + (1-\alpha) u_{t-1}, \quad \alpha = 0.5.
\end{equation}

\noindent\textbf{Reward.} The reward is designed to to promote articulation progress and grasp stability under perturbations jointly. The reward (Table~\ref{tab:reward_oracle_function}) balances articulation progress and grasp stability. Pose regularization terms $r^{\text{pos}}_t$ and $r^{\text{quat}}_t$ penalize deviation from the initial pose. Task completion is encouraged through $r^{\text{goal}}_t$, minimizing error between $\theta^{\text{art}}_t$ and the target $\theta^{\text{target}}_{s_t}$, and $r^{\text{timer}}_t$, rewarding sustained achievement of articulation state $s_t$. Incremental shaping $r^{\text{inc}}_t$ provides bonuses as $\theta^{\text{art}}_t$ crosses thresholds in $\Theta_{s_t}$ where $\Theta_{s_t}$ is a set of predefined articulation angle thresholds corresponding to the commanded state $s_t$. Stability is enforced via the contact term $r^{\text{contact}}_t$, penalizing deviation between the number of finger link contacts $n_t$ and the desired count $n^\star$, and the slippage term $r^{\text{slip}}_t$, activated when object height $h^{\text{obj}}_t$ drops below $h_{\min}$. Finally, the action penalty $r^{\text{act}}_t$ regularizes joint motion to prevent excessive excursions and over-tightening that may destabilize articulation.

\begin{table}[t]
\centering
\begin{tabular}{lll}
\toprule
Equation & Scale \\
\midrule
$r^{\text{pos}}_t = -\| x^{\text{obj}}_t - x^{\text{obj}}_0 \|_2^2$ & 500.0 \\
$r^{\text{quat}}_t = -\| q^{\text{obj}}_t - q^{\text{obj}}_0 \|_2^2$ & 5.0 \\
$r^{\text{goal}}_t = -\big| \theta^{\text{art}}_t - \theta^{\text{target}}_{s_t} \big|$ & 10.0 \\
$r^{\text{timer}}_t =
\begin{cases}
T^{\text{open}}_{t-1}\!+\!1, & s_t=1 \wedge \theta^{\text{art}}_t \geq \theta_{\text{open}}, \\
T^{\text{close}}_{t-1}\!+\!1, & s_t=0 \wedge \theta^{\text{art}}_t \leq \theta_{\text{close}}, \\
-1, & \text{otherwise}
\end{cases}$ & 0.05 \\
$r^{\text{inc}}_t = \sum_{\vartheta \in \Theta_{s_t}} w(\vartheta)\,\mathbf{1}\{\theta^{\text{art}}_t \ \text{crosses}\ \vartheta\}$ & 1.0 \\
$r^{\text{contact}}_t = n^\star - n_t$ & -0.1 \\
$r^{\text{slip}}_t = \mathbf{1}\{ h^{\text{obj}}_t < h_{\min} \}$ & -1.0 \\
$r^{\text{act}}_t = -\|u_t\|_2^2$ & -0.001 \\
\bottomrule
\end{tabular}
\caption{Reward function for the oracle policy.}
\label{tab:reward_oracle_function}
\vspace{-5mm}
\end{table}

\begin{figure}[t]
\centering

\begin{subfigure}[t]{0.58\columnwidth}
    \centering
    \includegraphics[width=\linewidth]{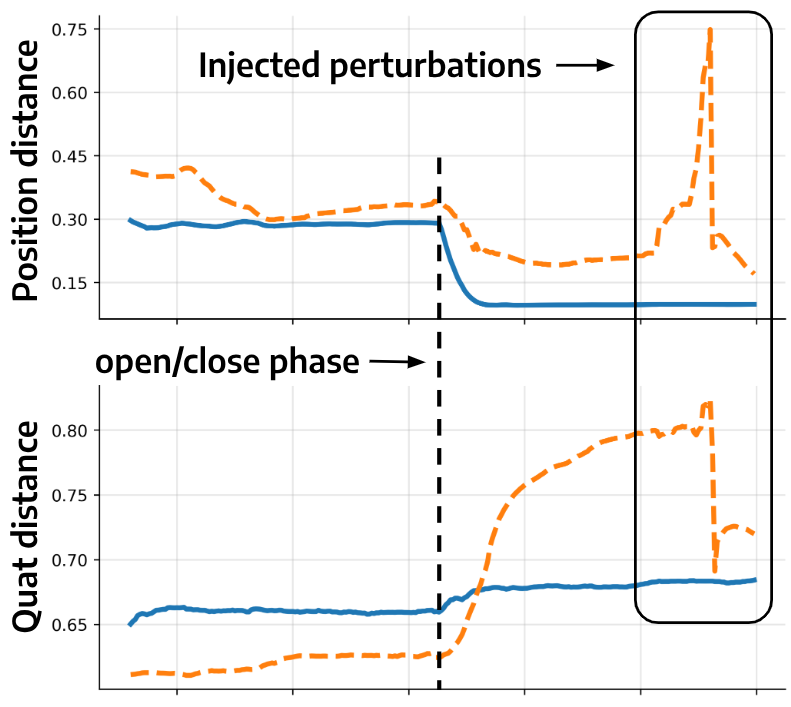}
    \caption{}
    % \caption{Motor and mimic joints. Motor torques $\tau^{\text{motor}}_t$ are computed from active joints only.}
    % \label{fig:motor_joints}
\end{subfigure}
\begin{subfigure}[t]{0.4\columnwidth}
    \centering
    \includegraphics[width=\linewidth]{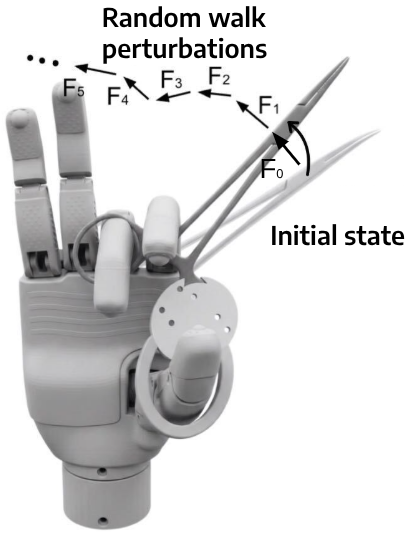}
    \caption{}
    % \caption{Augmented tactile structure with 3D-printed pad and foam layer.}
    % \label{fig:tactile_structure}
\end{subfigure}
\hfill

\caption{
(a) Tool pose variation during open and close phases under dynamic perturbations. 
The blue curve corresponds to the policy trained with injected perturbations (randomly sampled forces and torques), showing reduced deviation compared to the baseline. 
(b) Perturbation injection on the scissor, where forces and torques ($F_0, F_1 \dots$) are sequentially applied during articulation. Both forces and torques are sampled from a predefined ranges and updated as a random walks during execution.
}
\label{fig:perturb_policy}
\end{figure}

\noindent\textbf{Policy optimization with random walk perturbations.}
\label{sec:perturbations}
To further improve robustness beyond reward shaping alone, we introduce structured disturbance augmentation during training. External disturbances are simulated as a random walk during training. At each step, external forces $F^{\text{ext}}_t \in \mathbb{R}^3$ and torques $\tau^{\text{ext}}_t \in \mathbb{R}^3$ applied to the hand and tool are updated as
\begin{equation}
\begin{aligned}
F^{\text{ext}}_t &= \operatorname{clip}\!\left(F^{\text{ext}}_{t-1} + \Delta F_t^{\text{ext}},\, F_{\min}, F_{\max}\right), \\
\tau^{\text{ext}}_t &= \operatorname{clip}\!\left(\tau^{\text{ext}}_{t-1} + \Delta \tau_t^{\text{ext}},\, \tau_{\min}, \tau_{\max}\right),
\end{aligned}
\label{eq:random_walk_force_torque}
\end{equation}
where $\Delta F_t^{\text{ext}}$ and $\Delta \tau_t^{\text{ext}}$ are sampled from uniform ranges and clipped within predefined bounds. A random walk in force–torque space allows directional accumulation over time, covering disturbances that emulate gravity, acceleration, and external contact within a bounded domain. As shown in Fig.~\ref{fig:perturb_policy}, policies trained with random-walk perturbations exhibit improved robustness to external disturbances and the model produces joint configurations that more stably secure the articulated tool in both open and closed states.

\subsection{Base Policy Distillation}
While the oracle benefits from privileged state information, such signals are unavailable on hardware. We therefore distill the oracle into a deployable student policy that relies only on observable proprioceptive inputs. Standard distillation methods such as DAgger~\cite{ross2011reductionimitationlearningstructured} are ineffective for articulated in-hand manipulation, as partially informed students $\pi_{\text{student}}$ frequently drop objects early and fail to explore. Instead, we distill from stable oracle rollouts $\pi_{\text{oracle}}$ to train a proprioceptive student that operates without privileged inputs. Although visuotactile policies~\cite{qi2023generalinhandobjectrotation} achieve strong performance in simulation, they exhibit a significant sim-to-real gap due to occlusions and sensing discrepancies (see Sec.~\ref{sec:hardware_sensing}). In contrast, proprioceptive inputs $(q_t, s_t)$ are consistent in simulation and directly observable on hardware, making them more suitable for transfer. Accordingly, we first train an oracle policy $\pi_{\text{oracle}}$ using privileged state information in simulation, then distill its actions into a student policy $\pi_{\text{student}}$ that operates without privileged inputs. Random-walk force–torque perturbations (Eq.~\ref{eq:random_walk_force_torque}) are retained during training, allowing the student to develop robustness to gravity and contact disturbances under quasi-static manipulation.

% \begin{figure}
% \centering
%     \includegraphics[height=0.25\textwidth]{images/random_walk.png}
%     \caption{PositionPose deviation (top) and quaternion distance $\|q_t - q_0\|_2$ (bottom) of the articulated tool (surgical clamp).%     %  The quaternion metric represents orientation difference from the reference pose and should not be interpreted as quaternion magnitude.%      Blue denotes the policy trained with random-walk perturbation optimization (b), while orange denotes the policy trained without perturbations (a). The perturbation-trained policy maintains more stable multi-point contact with the tool, improving force closure and reducing torque-induced rotation under disturbances, resulting in lower pose deviation.}
%     \label{fig:perturb_policy}

% \end{figure}

\subsection{Online Adaptation via Cross-Attention Tactile Force Module (CATFA)}
\label{sec:hardware_sensing}

Although the distilled student policy enables sim-to-real transfer, it remains open-loop and lacks real-time sensory feedback. We therefore introduce an online adaptation mechanism to compensate for unmodeled contact dynamics. The distilled base policy $\pi_{\text{student}}$ provides a motion prior but operates open-loop, lacking real-time feedback. This is particularly limiting for articulated tools, where internal joint coupling and contact dynamics increase the likelihood of slip during in-hand execution. Although domain randomization improves simulation robustness, the absence of sensing prevents corrective responses to object slip and disturbances. To address this, we introduce \emph{Cross-Attention Tactile Force Adaptation (CATFA)}, which incorporates real tactile and joint force inputs to refine the behavior of $\pi_{\text{student}}$ and enable disturbance rejection. Unlike standard multimodal fusion that concatenates sensor features, CATFA treats the base policy embedding as a query and attends to sensor-derived keys and values. This intent-conditioned design enables targeted correction rather than symmetric feature aggregation, effectively serving as a learned impedance adaptation layer that injects feedback only when contact discrepancies arise.

\begin{figure}
\centering
    \includegraphics[width=0.5\textwidth]{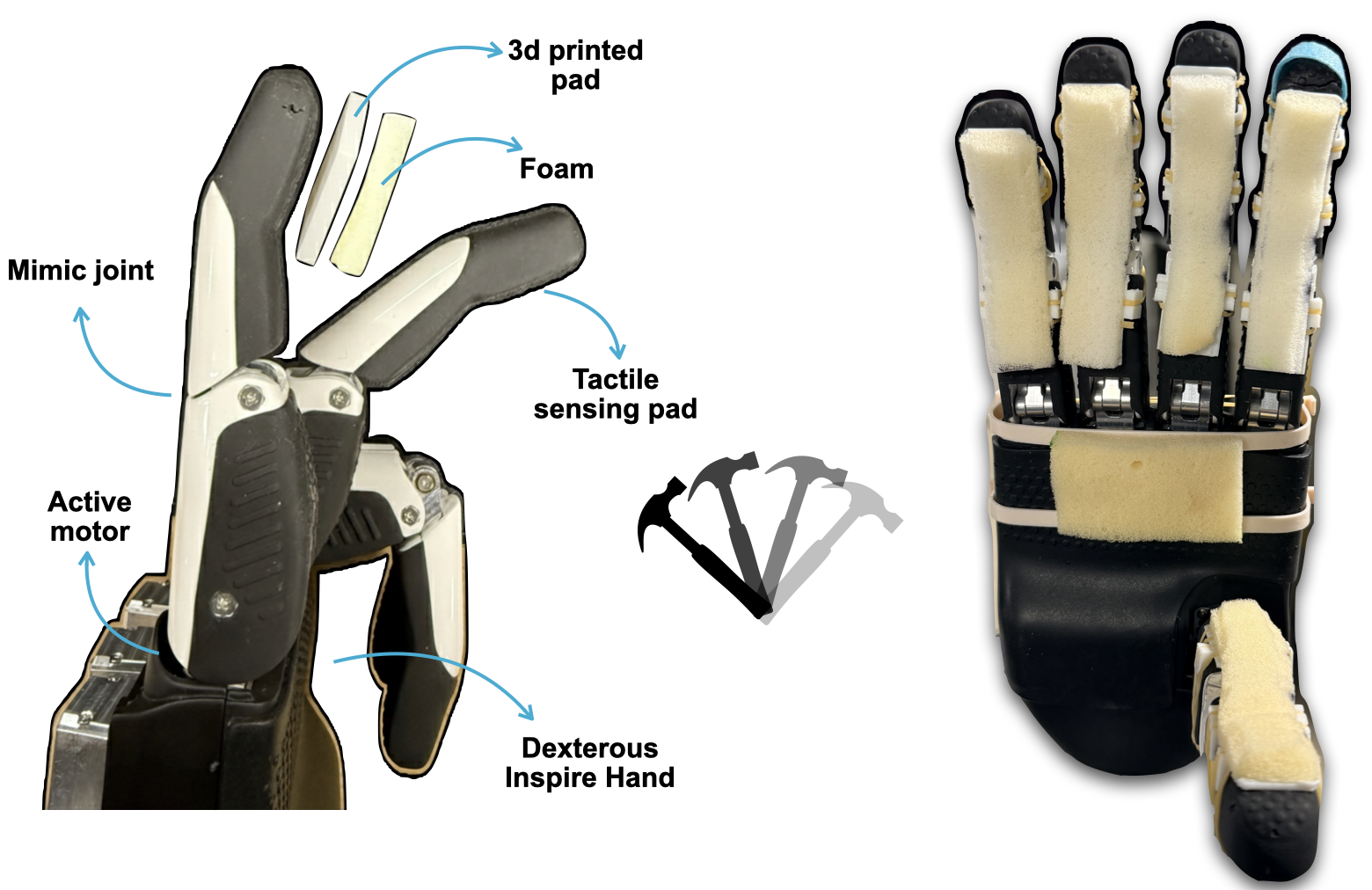}
    \caption{
        Inspire hand with an augmented tactile structure. A 3D-printed pad and foam layer redistribute contact loads to enhance tactile sensitivity and repeatability, while motor and mimic joints are illustrated for clarity. Motor torques $\tau^{\text{motor}}_t$ are computed only from active joints. These tactile and motor signals provide the hardware feedback used by CATFA for cross-attentive tactile–force refinement.
        }
    \label{fig:tactile_hardware}
    \vspace{-5mm}
\end{figure}

\noindent\textbf{Reducing Partial Observability via Tactile--Force Feedback.}
In simulation, the oracle observation $o_t$ provides a sufficient description of the articulated hand--tool state and induces a Markov process.
After distillation, the hardware policy observes only
\vspace{-3mm}
\[
s_t^{\text{prop}} = (q_t, s_t),
\]
which is a projection of the underlying physical state
\[
\chi_t =
(q_t,\,
\theta_t^{\text{art}},\,
x_t^{\text{obj}},\,
\eta_t),
\]
where $\eta_t$ denotes latent compliance and frictional effects.

\textit{Assumption (Load-Bearing Contact).}
During manipulation, the tool is fully supported by the hand; gravity and articulation torques are transmitted through finger contacts, yielding informative tactile and motor-torque signals. 
Augmenting the observation with
\[
y_t = (f_t^{\text{tact}}, \tau_t^{\text{motor}}),
\quad
s_t = (q_t, s_t, y_t),
\]
we model measurements as

\vspace{-3mm}
\[
y_t = h(\chi_t) + \epsilon, 
\qquad \epsilon \sim \mathcal{N}(0,\Sigma).
\]

Under this assumption, $y_t$ depends on $\eta_t$ through $h(\chi_t)$, implying
\[
I(\eta_t; y_t \mid q_t, s_t) > 0,
\]

\vspace{-1mm}
and therefore
\vspace{-1mm}
\[
H(\eta_t \mid q_t, s_t, y_t)
<
H(\eta_t \mid q_t, s_t).
\]
Thus tactile--force feedback reduces uncertainty over latent contact dynamics and moves the representation closer to Markovian.

\textit{Action-Space Correction.}
Let the optimal action under full state information be $u_t^* = \pi^*(\chi_t)$.
Under partial observability, the Bayes-optimal action is
\[
u_t^{\text{opt}} 
= 
\mathbb{E}_{\eta_t \mid q_t, s_t, y_t}
\big[ \pi^*(\chi_t) \big],
\]
which generally differs from the distilled base policy
\[
u_t^{\text{base}} = \pi_{student}(q_t, s_t).
\]
CATFA therefore learns a residual correction
\[
\hat{u}_t = u_t^{\text{base}} + \Delta u_t(q_t, s_t, y_t),
\]
where $\Delta u_t$ is trained from real rollouts to compensate for systematic discrepancies induced by latent contact dynamics.
While no explicit posterior over $\eta_t$ is computed, the correction implicitly captures the conditional dependence of the optimal action on tactile--force measurements.

\noindent\textbf{Hardware Sensing.} On the hardware side, additional tactile and motor force signals are available but are not accurately reproduced in simulation. Each finger is equipped with resistive tactile skins that provide quantized force readings $f^{\text{tact}}_t \in \mathbb{R}^{36 \times 44}$ over the pads. To improve sensitivity and repeatability under noisy sensing, we introduce a foam-plate-pad layer (Fig.~\ref{fig:tactile_hardware}) that redistributes local contact loads across neighboring taxels. We also measure motor torques $\tau^{\text{motor}}_t \in \mathbb{R}^{6}$ via current sensing. Geared actuators drive flexion, while extension relies on passive spring return. In simulation, the oracle policy $\pi_{\text{oracle}}$ uses actuator estimates $\tau^{\text{raw}}_t$, which differ from hardware measurements $\tau^{\text{motor}}_t$ due to transmission effects (e.g., gear ratios, friction, backlash, compliance, and sensor scaling). Modeling these effects would require detailed system identification; instead, we directly incorporate 
$f^{\text{tact}}_t$ and $\tau^{\text{motor}}_t$ into CATFA.

\noindent\textbf{Model Architecture.}
We now describe how these sensory modalities are integrated with the base policy. CATFA augments the frozen base policy $\pi_{\text{student}}$ with a sensor-driven refinement network. At each timestep $t$, the base policy $\pi_{\text{student}}$ produces an internal intent embedding $z_t \in \mathbb{R}^{64}$ from its final MLP layer. In parallel, the $f^{\text{tact}}_t \in \mathbb{R}^{36 \times 44}$ is processed by a convolutional encoder to yield a compact feature vector $\phi^{\text{tact}}_t \in \mathbb{R}^{64}$. In contrast,  $\tau^{\text{motor}}_t$ are mapped through a two-layer MLP to produce $\phi^{\text{force}}_t \in \mathbb{R}^{64}$. These sensor encodings are concatenated into $F_t = [\,\phi^{\text{tact}}_t, \phi^{\text{force}}_t\,] \in \mathbb{R}^{2 \times 64}$. A multi-head cross-attention module ($\text{MHA}$, with 8 heads and embed dimension 64) fuses intent and sensor signals by querying with $z_t$ against $F_t$.
\begin{equation}
    u_t = \text{MHA}\big(z_t,\, F_t,\, F_t\big) \in \mathbb{R}^{64}
\end{equation}
This formulation enables intent-conditioned feedback, where corrective signals are applied selectively based on the policy’s internal action embedding rather than through symmetric feature fusion. As attention applies corrections only when supported by sensory evidence, articulation behavior learned in simulation is preserved while allowing targeted adjustments. The adaptor adds minimal parameters and runs at the same control frequency without increasing inference latency.

\noindent\textbf{Training Adaptation.}  
We fine-tune the system by deploying the sim-to-real base policy $\pi_{\text{student}}$ on hardware under random disturbances and collecting fewer than 50 successful rollouts annotated via human feedback. Each rollout spans 2000 timesteps and includes multiple open/close phases, with the articulation command $s_t$ switching at random intervals (e.g., $t=50$ to $t=250$). This exposes the policy to diverse articulation states and transitions under gravity and contact effects. CATFA is trained via behavior cloning on a small human-labeled successful real-world rollouts. Given demonstrations $\mathcal{D}=\{(q_t, s_t, u_t)\}$, with oracle actions $u_t$, the student minimizes

\vspace{-4mm}
\begin{equation}
\mathcal{L}_{\text{BC}} = \mathbb{E}_{(q_t,s_t,u_t)\sim\mathcal{D}}
\big[\|\,u_t -\pi_{\text{student}}(q_t, s_t)\,\|_2^2\big]
\end{equation}
enabling efficient sim-to-real adaptation under contact-rich dynamics. During adaptation, the base policy $\pi_{\text{student}}$ remains frozen, allowing CATFA to inherit the motion prior while learning real-time corrective feedback. Together, the disturbance-trained oracle, distilled base policy, and cross-attention refinement module form a structured sim-to-real pipeline for articulated in-hand manipulation.

\noindent\textbf{Cross-Attention vs Concatenation:}
Direct concatenation of sensory embeddings perturbs the articulation prior symmetrically, regardless of whether contact discrepancies are present.
In contrast, cross-attention conditions corrections on the policy’s internal intent representation $z_t$, enabling targeted feedback only when tactile or torque signals indicate deviation from expected contact behavior.
This intent-conditioned design preserves nominal articulation trajectories while selectively compensating for unmodeled contact dynamics.

\vspace{-2mm}
\section{Experiments}
% \subsection{Evaluation Metrics}
% For Tab \ref{tab:perturbation_all}, we report orientation deviation using the $\ell_2$ distance between unit quaternions, $\|\mathbf{q}_t - \mathbf{q}_0\|_2$, which measures relative orientation error rather than quaternion magnitude; for small angular deviations, this metric is locally proportional to the geodesic quaternion distance and preserves qualitative trends. 
% To quantify in-hand stability, we define pose deviation as the combined translational and rotational error,
% \begin{equation}
% e_{\text{pose}}(t) =
% \left\| \mathbf{x}^{\text{obj}}_{t} - \mathbf{x}^{\text{obj}}_{0} \right\|_{2}
% + \lambda
% \left\| \mathbf{q}^{\text{obj}}_{t} - \mathbf{q}^{\text{obj}}_{0} \right\|_{2},
% \end{equation}
% where $\mathbf{x}^{\text{obj}}_{t}$ and $\mathbf{q}^{\text{obj}}_{t}$ denote object position and unit quaternion orientation, and $\lambda$ balances translational and rotational contributions. 
% In Tab \ref{tab:results_art_manip}, articulation performance is evaluated via success rate (maintenance of articulation angle thresholds over the episode), maximum tip separation in the open state, and residual tip gap in the closed state. 
% Robustness under perturbations is quantified using $e_{\text{pose}}(t)$, and all results are reported as mean $\pm$ standard deviation over $n=10$ independent real-world rollouts per tool.

\begin{figure}[t]
    \centering
    \includegraphics[width=0.49\textwidth]{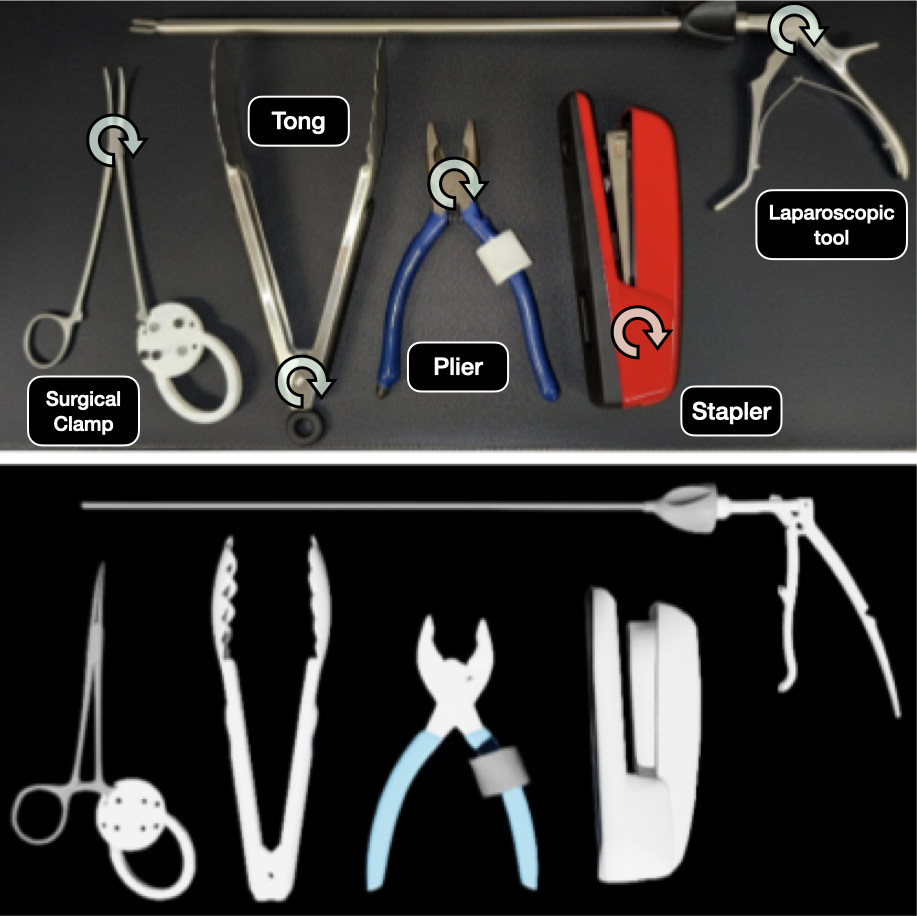}
    \caption{Articulated tools used in experiments: real-world examples (top) and simulated counterparts (bottom). The axis denotes the articulation rotation axis. These tools span a wide range of articulation types and gripping strategies, capturing diverse kinematic structures and manipulation provisions.}
    \label{fig:object_set}
    \vspace{-7mm}
\end{figure}

\textbf{Articulated Tool Set.}
We evaluate on five articulated tools, each modeled with a single revolute joint and instantiated in both simulation and hardware (Fig.~\ref{fig:object_set}). In simulation, object mass, surface friction, and material properties are randomized to improve transfer robustness. While simulated joints capture intended kinematics, they do not fully reproduce real-world mechanics, introducing a domain gap. To account for scale differences, mechanical fixtures are attached so the tools fit the Inspire hand, which is approximately 1.4× the size of an average human hand. This provides a controlled yet challenging benchmark for articulated in-hand manipulation.

\textbf{Hardware.}
We use the Inspire Hand with 6 active DoFs (one per finger, two for the thumb) and 6 mimic joints. Joint targets are issued at 30~Hz and tracked by a low-level PD controller at 120~Hz. Tactile sensing operates at 75~Hz using a $36 \times 44$ resistive array, with additional force measurements at all active joints. The hand is covered with a nitrile glove for consistent contact properties. Initial grasp configurations are collected using Manus MetaGloves Pro~\cite{manus2025} (Sec.~\ref{sec:oracle_policy}), and tactile augmentation details are provided in Sec.~\ref{sec:hardware_sensing} and Fig.~\ref{fig:tactile_hardware}.

\textbf{Simulation.}
Policies are trained in IsaacLab~\cite{mittal2023orbit}. Each environment contains a hand and an articulated tool initialized to a feasible state. The simulator runs at 120 Hz with control decimation of 3 (effective 40 Hz). Episodes last 2000 steps ($\approx$50s) and terminate upon object drop, workspace exit, or horizon completion. Contact dynamics are approximated using convex decomposition for complex geometries and convex hulls for simpler ones. We run 8192 parallel headless environments on a single Nvidia RTX 4090. Training for each tool requires 4–12 hours, and more complex collision models increase the computational cost.

\begin{figure*}[t]
    \centering
    % --- Left panel ---
    \begin{subfigure}[b]{0.49\textwidth}
        \centering
        \includegraphics[width=\textwidth]{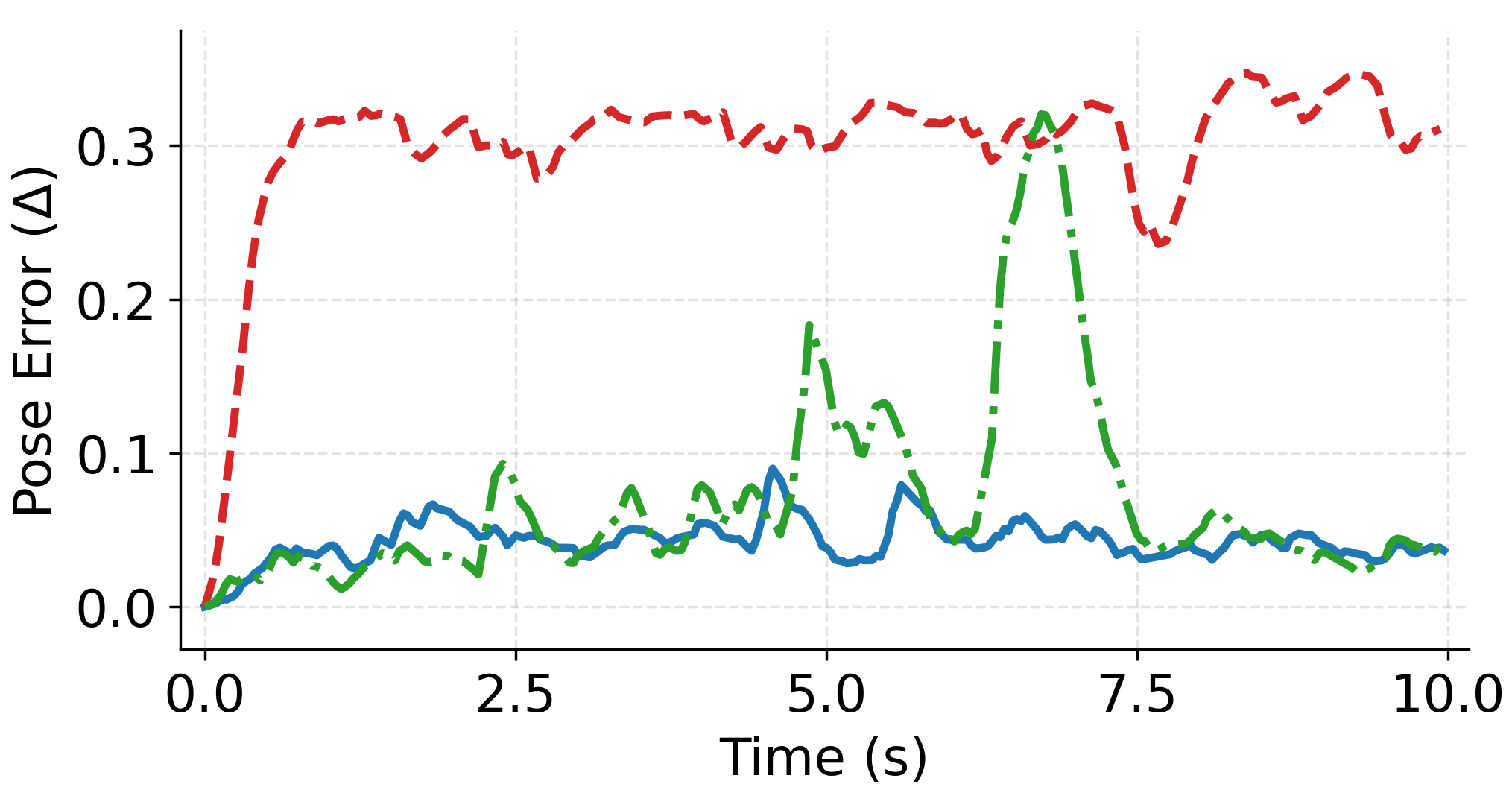}
        \caption{Closed trajectory comparison}
        \label{fig:traj_metric_close}
    \end{subfigure}
    \hfill
    % --- Right panel ---
    \begin{subfigure}[b]{0.49\textwidth}
        \centering
        \includegraphics[width=\textwidth]{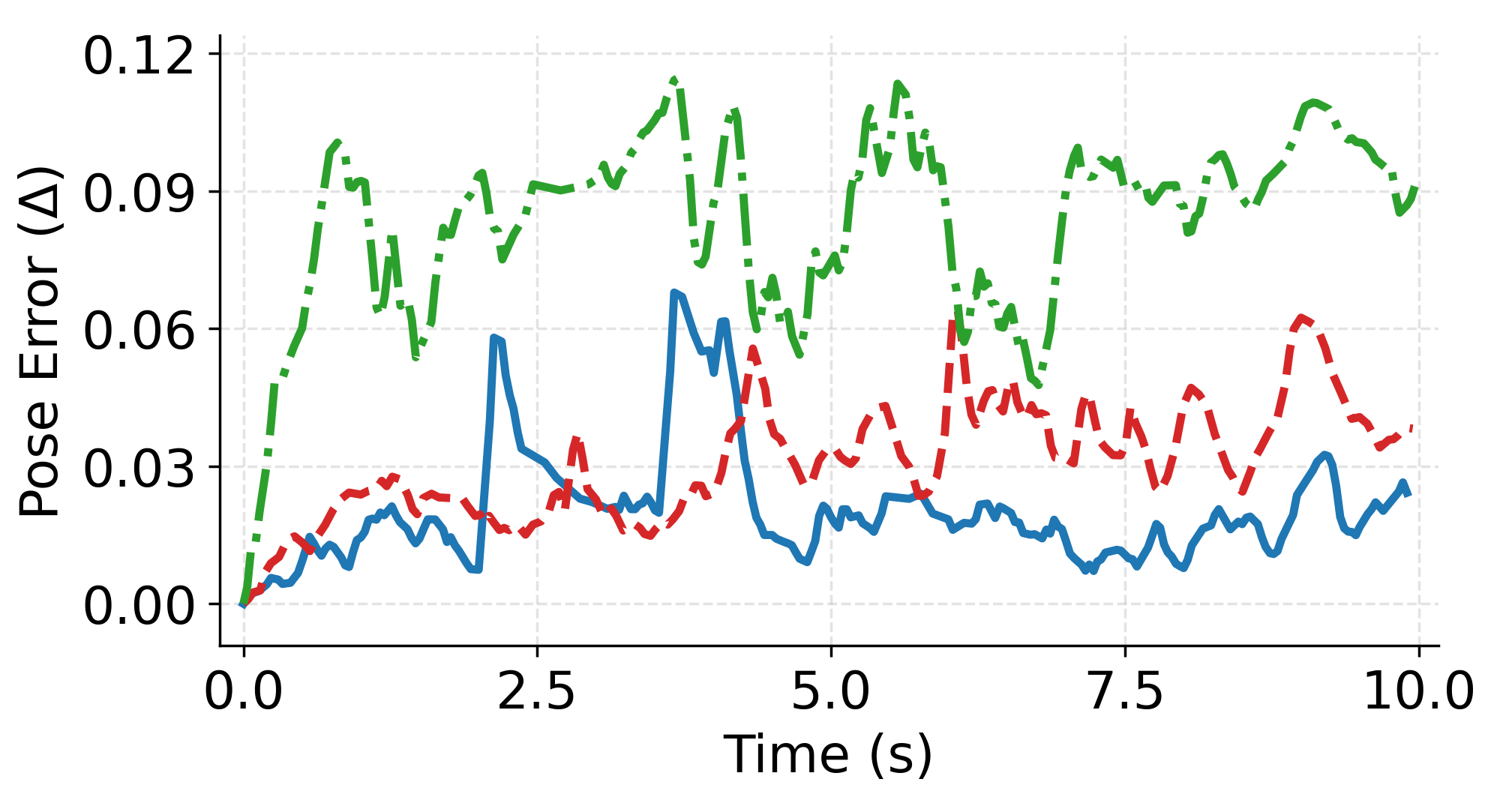}
        \caption{Open trajectory comparison}
        \label{fig:traj_metric_open}
    \end{subfigure}

    \caption{Closed-loop disturbance response during articulated manipulation.
    We plot the composite pose deviation $\Delta(t) = \|x^{obj}_t - x^{obj}_0\|_2 + \lambda \|q^{obj}_t - q^{obj}_0\|_2$. (a) Closed articulation phase. (b) Open articulation phase. CATFA (solid blue) yields a uniformly lower mean error and attenuates high-frequency oscillations induced by disturbances. The proprioceptive BC policy (dashed red) accumulates drift, while the distilled student (dash-dotted green) exhibits large transient error spikes under contact perturbations. These results indicate improved closed-loop robustness and sensor-conditioned stabilization of internal articulation dynamics.}
    \label{fig:traj_metric}
\end{figure*}

\begin{table}[t]
\centering
\setlength{\tabcolsep}{3.5pt}
\renewcommand{\arraystretch}{0.95}
\resizebox{\columnwidth}{!}{%
\begin{tabular}{ll|c|c|c}
\toprule
\textbf{Policy} & \textbf{Articulated Tool} 
& \textbf{Succ (\%) $\uparrow$} 
& \textbf{Opening Disp. (mm) $\uparrow$} 
& \textbf{Closure Residual (mm) $\downarrow$} \\
\midrule

\multicolumn{5}{l}{\textit{Simulation (Oracle, reference only)}} \\
 & Surgical Clamp   & 100 & 17.26 $\pm$ 0.63   & 0.0 $\pm$ 0.0 \\
 & Tong             & 100 & 51.55 $\pm$ 2.93   & 17.23 $\pm$ 1.45 \\
 & Plier            & 100 & 17.45 $\pm$ 1.20   & 9.34 $\pm$ 2.10 \\
 & Laparoscopic Tool& 100 & 99.31 $\pm$ 2.24   & 67.62 $\pm$ 1.58 \\
 & Stapler          & 100 & 22.88 $\pm$ 1.44   & 0.000 $\pm$ 0.000 \\
\midrule
\midrule

\multicolumn{5}{l}{\textit{Sim-to-Real (Student)}} \\
 & Surgical Clamp   & 20  & 6.8 $\pm$ 2.3    & 3.4 $\pm$ 3.8 \\
 & Tong             & 100 & 51.23 $\pm$ 3.97   & 29.17 $\pm$ 1.26 \\
 & Plier            & 30  & 9.81 $\pm$ 3.30    & 9.06 $\pm$ 9.36 \\
 & Laparoscopic Tool& 100 & \textbf{109.45 $\pm$ 9.98}  & 76.98 $\pm$ 4.52 \\
 & Stapler          & 100 & 11.32 $\pm$ 4.70    & 1.03 $\pm$ 0.97 \\
\midrule

\multicolumn{5}{l}{\textit{Proprioceptive BC}} \\
 & Surgical Clamp   & 90  & 6.6 $\pm$ 0.66    & 1.46 $\pm$ 0.45 \\
 & Tong             & 100 & 51.55 $\pm$ 0.87   & 32.64 $\pm$ 1.7 \\
 & Plier            & 70  & 7.73 $\pm$ 1.87    & 3.68 $\pm$ 0.21\\
 
 & Laparoscopic Tool& 100 & 107.39 $\pm$ 1.48  & 75.07 $\pm$ 2.06 \\
 & Stapler          & 100 & 13.08 $\pm$ 0.48    & 0.6 $\pm$ 0.5 \\
\midrule

\multicolumn{5}{l}{\textbf{CATFA (Ours)}} \\
 & Surgical Clamp   & 100 & \textbf{8.5 $\pm$ 0.8}     & \textbf{0.0 $\pm$ 0.0} \\
 & Tong             & 100 & \textbf{64.23 $\pm$ 0.36}   & \textbf{28.31 $\pm$ 0.91} \\
 & Plier            & 100 & \textbf{11.28 $\pm$ 0.76}   & \textbf{3.12 $\pm$ 0.16} \\
 & Laparoscopic Tool& 100 & 109.31 $\pm$ 1.05  & \textbf{73.28 $\pm$ 1.6} \\
 & Stapler          & 100 & \textbf{14.52 $\pm$ 0.59}   & \textbf{0.15 $\pm$ 0.09} \\
\bottomrule
\end{tabular}%
}
\caption{Quantitative evaluation on articulated tool manipulation. 
Simulation (Oracle) results are shown for reference only and are not considered for bold comparison. 
\textbf{Opening Disp.} denotes maximum achieved tip separation in the open state (higher is better). 
\textbf{Closure Residual} denotes the remaining tip gap in the closed state (lower is better). 
Best real-world results per tool are highlighted in bold.}
\label{tab:results_art_manip}
\vspace{-5mm}
\end{table}

% \vspace{-6mm}
\subsection{Baselines and Ablation Study}
% We perform two sets of ablation studies to evaluate the contribution of tactile and force feedback. In the first study as shown in Tab. \ref{tab:results_art_manip}, we compare (i) the sim-to-real base policy $\pi_{\text{student}}$, (ii) the CATFA-augmented policy, and (iii) a behavior-cloned policy using only proprioceptive inputs (no tactile or force signals). Performance is measured on approximately 10 real-world rollouts per object by recording the articulation joint angle after each open/close phase and averaging across trials. 

We perform two experiments to evaluate the role of tactile and force feedback. 
In the first study (Tab.~\ref{tab:results_art_manip}), we compare the sim-to-real student policy $\pi_{\text{student}}$, a Proprioceptive BC policy (without tactile or force inputs), and CATFA. 
For each articulated tool, we conduct approximately 10 real-world rollouts and evaluate articulation performance using three metrics: success rate, maximum tip separation in the open state, and residual tip gap in the closed state. 
Success is defined as maintaining the commanded open or closed configuration (joint angle above or below a predefined threshold) throughout the episode, and tip metrics are computed from endpoint separation after each open/close phase and averaged across trials.

In the second study (Tab.~\ref{tab:perturbation_all}), we mount the dexterous hand on a Franka arm and execute predefined end-effector trajectories with randomized accelerations to induce motion-planning disturbances. 
In addition to the above baselines, we include Proprio–Force BC, Proprio–Tactile BC, and Proprio–Tactile–Force BC, where tactile and/or force embeddings are directly concatenated without cross-attention refinement. 
We track the articulated tool pose using ArUco markers and quantify robustness via pose deviation,
\vspace{-2mm}
\begin{equation}
e_{\text{pose}}(t) =
\left\| \mathbf{x}^{\text{obj}}_{t} - \mathbf{x}^{\text{obj}}_{0} \right\|_{2}
+ \lambda
\left\| \mathbf{q}^{\text{obj}}_{t} - \mathbf{q}^{\text{obj}}_{0} \right\|_{2},
\end{equation}
\vspace{-0.5mm}
where $\mathbf{x}^{\text{obj}}_{t}$ and $\mathbf{q}^{\text{obj}}_{t}$ denote object position and unit quaternion orientation, and $\lambda$ balances translational and rotational contributions. 
Orientation deviation is computed using the $\ell_2$ distance between unit quaternions, $\|\mathbf{q}_t - \mathbf{q}_0\|_2$, which approximates the geodesic distance for small angular deviations. 
All results are reported as mean $\pm$ standard deviation over $n=10$ independent real-world rollouts per tool.

% \subsection{Baselines and Evaluation}

% We evaluate on a Franka arm with a dexterous hand. Baselines include:  
% \begin{itemize}  
%   \item Base student policy (no adaptation),  
%   \item Behavior cloning using only proprioception,  
%   \item Sim-to-real with domain randomization.  
% \end{itemize}

% \subsection{Robot setup}

% \begin{figure}
%     \centering
%     \includegraphics[width=0.35\textwidth]{images/tactile.png}
%     \caption{New tactile setup for sensitivity}
%     \label{fig:tactile}
% \end{figure}
% \begin{itemize}
%     \item Inspire hands setup
%     \item Manus gloves for initial pose setup
%     \item Simulation digital clone
%     \item On policy training for articulated tool manipulation
%     \item Tactile sensing setup
%     \item Off-policy training for adapting tactile feedback
% \end{itemize}

% \subsubsection{Object Sets}

% \subsubsection{Simulation Setup}
% \begin{itemize}
%     \item Our reinforcement learning policy was trained in IsaacLab to simulate the manipulation of articulated tools with the Inspire robotic hand. We selected IsaacLab for its significant advantages over the preceding IsaacGym framework.
%     \item All parallelized training instances were executed on a single NVIDIA 4090 GPU, which provided the necessary computational power for the complex simulations.
% \end{itemize}
\vspace{-2mm}
\section{Results}

\begin{table*}[t]
\centering
\footnotesize
\setlength{\tabcolsep}{3.5pt} % tighter columns
\renewcommand{\arraystretch}{1.05} % slightly taller rows
\resizebox{\textwidth}{!}{%
\begin{tabular}{l|c|c||c|c||c|c}
\toprule
\multirow{2}{*}{Policy} &
\multicolumn{2}{c}{Clamp} &
\multicolumn{2}{c}{Plier} &
\multicolumn{2}{c}{Laproscopic tool} \\
\cmidrule(lr){2-3}\cmidrule(lr){4-5}\cmidrule(lr){6-7}
& Open (m) & Close (m) & Open (m) & Close (m) & Open (m) & Close (m)\\
\midrule
Sim-to-Real (Student) &
\makecell{0.056 $\pm$ 0.04} &
\makecell{0.083 $\pm$ 0.038} & \makecell{0.265 $\pm$ 0.078} & \makecell{0.188 $\pm$ 0.068} & \makecell{0.242 $\pm$ 0.135} & \makecell{0.09 $\pm$ 0.044} \\
Proprioception BC &
\makecell{0.083 $\pm$ 0.02} &
\makecell{0.053 $\pm$ 0.017} & \makecell{0.096 $\pm$ 0.084} & \makecell{0.073 $\pm$ 0.045} & \makecell{0.159 $\pm$ 0.096} & \makecell{0.053 $\pm$ 0.067} \\
Proprio–Tactile-Force BC &
\makecell{0.031 $\pm$ 0.035} &
\makecell{0.046 $\pm$ 0.068} & \makecell{0.046 $\pm$ 0.090} & \makecell{\textbf{0.031 $\pm$ 0.055}} & \makecell{0.87 $\pm$ 0.073}& \makecell{0.049$\pm$ 0.053}\\
Proprio–Tactile BC &
\makecell{0.049 $\pm$ 0.04} &
\makecell{0.045 $\pm$ 0.039} & \makecell{0.051 $\pm$ 0.043} & \makecell{0.065 $\pm$ 0.042} & \makecell{0.89 $\pm$ 0.067}& \makecell{0.052$\pm$ 0.060}\\
Proprio–Force BC &
\makecell{0.046 $\pm$ 0.038} &
\makecell{0.046 $\pm$ 0.036} & \makecell{0.051 $\pm$ 0.042} & \makecell{0.056 $\pm$ 0.053} & \makecell{0.59 $\pm$ 0.058}& \makecell{0.076$\pm$ 0.066}\\
\textbf{CATFA (Ours)} &
\makecell{\textbf{0.022 $\pm$ 0.018}} &
\makecell{\textbf{0.035 $\pm$ 0.024}} & \makecell{\textbf{0.039 $\pm$ 0.026}} & \makecell{0.033 $\pm$ 0.018} & \makecell{\textbf{0.05 $\pm$ 0.066}}& \makecell{\textbf{0.044$\pm$ 0.086}}\\
\bottomrule
\end{tabular}}
\caption{
\textbf{Perturbation analysis} across three articulated tools in open and closed states.
Each entry reports mean $\pm$ std of pose deviation $e_{\text{pose}}$ (m) $\downarrow$, where
$e_{\text{pose}}(t)=\|\mathbf{x}_t-\mathbf{x}_0\|_2+\lambda\|\mathbf{q}_t-\mathbf{q}_0\|_2$,
averaged over $n=10$ real-world rollouts while the Franka executes a predefined random trajectory with injected perturbations.
Lower is better. m --> meters
}
\label{tab:perturbation_all}
\vspace{-5mm}
\end{table*}

\label{sec:results}
An overview of the pipeline is shown in Fig.~\ref{fig:overview}. 
We evaluate the effectiveness of CATFA for sim-to-real adaptation against multiple behavior-cloning baselines.

\textbf{Baselines.} 
For articulated manipulation (Tab.~\ref{tab:results_art_manip}), we compare CATFA with (1) a Proprioceptive BC policy (no tactile or force inputs) and (2) the distilled sim-to-real policy $\pi_{\text{student}}$ deployed on hardware with limited fine-tuning.

\textbf{Real-world evaluation.} 
We evaluate on five articulated tools (Fig.~\ref{fig:object_set}). 
Quantitative results are reported on in-domain objects, with additional qualitative demonstrations on tools varying in weight and geometry. 
Success is defined by maintaining articulation angle thresholds across repeated open–close cycles (Tab.~\ref{tab:results_art_manip}). 
CATFA achieves the most consistent articulation, reducing variance and failure under perturbations compared to all baselines.

\textbf{Robustness to perturbations.} 
For the dynamic evaluation (Tab.~\ref{tab:perturbation_all}), we additionally compare Proprio–Force BC, Proprio–Tactile BC, and Proprio–Tactile–Force BC, where tactile and/or force embeddings are directly concatenated without cross-attention refinement. 
We execute a predefined manipulation trajectory on the Franka hand–tool setup with randomized accelerations and external disturbances, and measure pose error combining position and quaternion deviations. 
CATFA yields the lowest or near-lowest error across tools, demonstrating improved disturbance rejection over direct-embedding baselines and the sim-to-real student.

\textbf{Simulation validation.} 
In simulation, we compare policies trained with and without random-walk force–torque perturbations (Eq.~\ref{eq:random_walk_force_torque}). 
Perturbation-trained policies exhibit improved stability under disturbances, supporting the robustness gains observed on hardware.

\vspace{-2mm}
\section{Discussion \& Conclusion}

Our results validate a disturbance-driven sim-to-real framework for in-hand articulation. 
Structured grasp initialization aligns the hand with tool joints, while simulation enables learning of contact-rich skills that transfer reliably through proprioceptive policies.

Robustness is introduced at the simulation stage by training the privileged policy under structured force–torque random-walk perturbations. 
The distilled student inherits this disturbance-aware behavior without directly observing the perturbations. CATFA further refines the frozen student through an intent-conditioned cross-attention adaptor that fuses tactile and motor torque feedback to produce contact-aware corrections without retraining the base policy. Real-world experiments across five articulated tools confirm improved stability and disturbance rejection. This modular design supports extensibility: articulation policies serve as reusable skill primitives, and additional sensing modalities can be integrated via dedicated adaptors.

During hardware experiments, minor joint micro-oscillations are occasionally observed in the Inspire hand due to backlash in the geared transmission and spring compliance not modeled in the simulation. 
These oscillations are reduced under CATFA compared to the proprioceptive BC baseline and do not affect articulation success.

\noindent\textbf{Limitations and Future Work.}
The current formulation assumes binary articulation commands $s_t \in \{0,1\}$, limiting applicability to tools with discrete open--close behaviors.
Extending CATFA to continuous articulation targets and multi-DOF articulated mechanisms remains future work.
Hardware fine-tuning is sensitive to motor torque scaling and control-frequency discrepancies between simulation and the geared, spring-assisted hand.
Integrating adaptive torque calibration or online system identification may further reduce sim-to-real discrepancies arising from transmission effects.

\balance
\bibliographystyle{ieeetr}
\bibliography{root}

\newpage

\end{document}